\documentclass{article}
\usepackage{spconf,amsmath,amssymb,graphicx,hyperref}
\usepackage{booktabs}
\usepackage{algorithm}
\usepackage{algpseudocode}

\title{SOURCE-PRESERVING ALIGNMENT FOR ROBUST EVIDENCE LOCALIZATION IN SCIENTIFIC PDFS}

\name{
Zihao Liu$^{1,3}$,
Wei Yang$^{2,3}$,
Zixiao Dong$^{1,3}$,
Chenshu Li$^{1,3}$,
Longzhang Liu$^{1,3}$,
Tao Tan$^{4}$,
Hong Xie$^{1,3*}$
\thanks{
$*$ Corresponding author: Hong Xie (hongx87@ustc.edu.cn).
}
}

\address{
$^{1}$School of Computer Science and Technology, University of Science and Technology of China\\
$^{2}$University of Science and Technology of China, 
$^{3}$State Key Laboratory of Cognitive Intelligence,\\
$^{4}$CCCC Second Highway Consultants Co., Ltd.
}

\begin{document}
\maketitle

\begin{abstract}
Scientific information-extraction systems often return a claim with an evidence string, which users must locate in the original PDF. This is challenging because the extracted evidence and PDF text layer are different representations: line wrapping, Unicode variants, superscripts, citation markers, and fragmented items alter text sequences and geometry. We present a source-preserving alignment framework: normalize text for robust matching while preserving provenance for accurate localization. It aligns evidence with normalized page text, maps matches back to source-character spans, and renders only their geometry. When exact alignment fails, line-break-aware token alignment recovers supported spans while excluding unmatched noise. Experiments on 1,020 chemistry papers show that the framework achieves a 92.6\% quote-level automatic localization rate, compared with 43.6\% for text search and 19.1\% for a precomputed bounding-box baseline. Component ablation confirms distinct contributions from normalization and approximate token alignment, while human verification assesses the visual correctness of returned highlights. Overall, these results demonstrate that reliable evidence verification requires robust matching and precise localization within a shared source-preserving alignment representation.
\end{abstract}


\section{Introduction}
\label{sec:intro}
Retrieval-augmented generation~\cite{lewis2020rag} and scientific claim verification~\cite{wadden2020scifact} use retrieved evidence, while S2ORC provides structured scientific text~\cite{lo2020s2orc} and citation-generation research stresses that answers should be verifiable~\cite{gao2023alce}. Making evidence useful requires locating it on the original PDF page. Unlike standard text search, this task must match evidence from Markdown or other logical formats to PDF text with line and item breaks, different dash forms, citation markers, and separate superscripts. These differences arise from the PDF text model and character encoding~\cite{iso32000,unicode} and remain a challenge in scholarly-PDF extraction~\cite{lopez2009grobid,tkaczyk2015cermine}. Matched text must then be turned into precise page-level highlights. Plain text search is sensitive to normalization differences; precomputed bounding boxes cannot handle text not expected at indexing time and are sensitive to changes in layout and encoding. Neither preserves the character-level link to source geometry needed for accurate highlighting.

We frame evidence localization as alignment among normalized text for matching, source characters for provenance, and page geometry for rendering. Our principle, \emph{normalize for matching, preserve provenance for localization}, links each normalized character to its source item and offset through an explicit provenance map. Handling extraction differences alone cannot ensure accurate highlight boundaries. We contribute (1) a normalized representation that preserves these source links, (2) a method that uses it to turn exact or approximate text alignment into character spans and highlights based on page geometry, and (3) an evaluation on scientific chemistry PDFs against text-search and precomputed bounding-box baselines, showing higher localization success and more accurate highlight boundaries, especially for long evidence strings.

\section{Related Work}
\label{sec:related-work}

PDF libraries such as PDF.js~\cite{pdfjs} and PyMuPDF~\cite{pymupdf} expose page strings, text items, and geometry, while neural parsers improve structured extraction but do not guarantee traceable visual source spans~\cite{blecher2024nougat}. Standard search is fragile to extraction-order, encoding, and layout differences, whereas bounding-box indexes are limited by text-to-box correspondence.

This correspondence is particularly difficult for scientific PDFs, where equations, formulas, Greek letters, superscripts, and reference markers can be fragmented or nonstandard. Document-layout resources likewise show substantial visual and structural variation~\cite{zhong2019publaynet,li2020docbank,shen2021layoutparser,kim2022doclaynet}. A robust locator must therefore combine tolerant comparison with faithful recovery of source characters and geometry.

Approximate string matching handles textual errors~\cite{navarro2001}, while classical sequence alignment~\cite{needleman1970} and longest-common-subsequence (LCS) algorithms~\cite{hirscheberg1975} identify ordered correspondences despite gaps. Our token-level fallback uses LCS correspondences to recover supported tokens; PDF evidence localization additionally projects them through a provenance map to source characters and page geometry.

\section{Proposed Method}
\subsection{Deterministic Matching Procedure}
Let $q$ denote an evidence quote and let $P=\{p_1,\ldots,p_n\}$ denote the ordered PDF text items, where each $p_i$ contains source text $s_i$ and transform $T_i$. The method uses fixed rules rather than learned parameters. It first normalizes and searches for an exact occurrence:
\begin{equation}
 \hat q=N(q), \qquad \hat p=N(s_1)\mathbin{\|}\cdots\mathbin{\|}N(s_n).
\end{equation}

If exact matching fails, let $\mathbf q=(q_1,\ldots,q_m)$ and $\mathbf p$ be the normalized quote and page token sequences. For $m\geq5$, the method enumerates a fixed set of candidate windows and selects
\begin{equation}
 \tilde{\mathbf w}=
 \arg\max_{\mathbf w\in\mathcal{C}(\mathbf q,\mathbf p)}
 \mathrm{score}(\mathbf q,\mathbf w).
\end{equation}
The fallback result is returned only when
$\mathrm{score}(\mathbf q,\tilde{\mathbf w})\geq\tau$, where $\tau=0.8$; otherwise, no fallback match is returned. Thus, the procedure performs deterministic selection from a finite candidate set rather than parameter optimization.

\subsection{Normalization and Provenance Mapping}
We define a normalization function $N(\cdot)$ that applies Unicode compatibility normalization, dash/minus unification, malformed-encoding repair, scientific-symbol mapping, and removal of layout-only separators, consistent with the Unicode representation principles used by PDF text workflows \cite{unicode,iso32000}. The normalized quote and page stream are
\begin{equation}
 \hat q=N(q), \qquad \hat p=N(s_1)\mathbin{\|}N(s_2)\mathbin{\|}\cdots\mathbin{\|}N(s_n).
\end{equation}
During normalization, every output character stores a pair $(i,c)$ identifying its source text item $i$ and source character offset $c$. Thus, an exact match in $\hat p$ can be projected back to the corresponding source intervals.

The normalization includes compatibility rules for symbols commonly damaged during PDF extraction. These rules are used only for comparison; the original page text is never rewritten.

\subsection{Exact Matching and Character-Range Clipping}
If $\hat q$ occurs contiguously in $\hat p$, the method collects all source intervals touched by the match. For each text item, only the interval $[c_{\mathrm{start}},c_{\mathrm{end}})$ is rendered. Because PDF.js text items do not generally expose per-glyph rectangles, the horizontal extent is estimated proportionally from the item width:
\begin{equation}
 x_{\mathrm{start}}=x_0+w\frac{c_{\mathrm{start}}}{|s_i|}, \qquad
 x_{\mathrm{end}}=x_0+w\frac{c_{\mathrm{end}}}{|s_i|}.
\end{equation}
The resulting rectangle is transformed using the item's PDF matrix and clipped to the page viewport. This prevents a quote ending inside a text item from highlighting the next sentence contained by the same item.

\subsection{Line-Break and Noisy-Text Alignment}
When exact matching fails, both texts are tokenized. Alphabetic line-end hyphenation is merged when its fragments form a word, while formula-like strings such as ``ZSM-5'' are preserved. For a quote of $m$ tokens, each scale $\alpha\in\{0.70,0.85,1.00,1.15,1.30\}$ produces a candidate length $\ell_{\alpha}=\operatorname{round}(\alpha m)$, using nearest-integer rounding with half values rounded upward. Windows longer than the page-token sequence are omitted.

For a candidate token window $\mathbf w=(w_1,\ldots,w_{\ell})$, we use the longest common subsequence~\cite{hirscheberg1975} to define the score as
\begin{equation}
 \mathrm{score}(\mathbf q,\mathbf w)=
 \frac{\mathrm{LCS}(\mathbf q,\mathbf w)}
 {\max(|\mathbf q|,|\mathbf w|)}
 =
 \frac{\mathrm{LCS}(\mathbf q,\mathbf w)}{\max(m,\ell)}.
\end{equation}
Both the LCS count and its denominator are measured in tokens. The renderer highlights only LCS-aligned tokens, excluding unaligned markers.

\subsection{Geometric Rendering}
For each matched source interval, the method computes the four transformed corners of the text rectangle from the PDF.js transform matrix. Rectangles are clamped to normalized page bounds and merged only when they are nearby fragments on the same rendered line. This avoids merging across columns, headers, or unrelated text regions.

\section{Experiments and Analysis}
\label{sec:experiments-analysis}

\subsection{Dataset}

The corpus contains 1,020 publicly accessible chemistry papers paired one-to-one with Markdown evidence files, comprising 18,580 quote instances (18.2 per paper). All PDF--Markdown filename pairs were verified; no PDF was missing. Only papers with both files available were retained, so the benchmark evaluates localization rather than upstream claim extraction. Quotes vary in length, and PDFs include multi-column pages, chemical formulas, figure references, and noisy text layers, exposing ordinary search failures and boundary errors after apparently successful matches. For reproducibility, we provide the PDF--Markdown pairing rule, quote parser, BBOX records, and comparison scripts. The corpus manifest records each paper's identifier, public access source, access date, and redistribution status.

\subsection{Comparison Methods}

We compare three methods. Proposed uses source-preserving normalization, alignment, source-span recovery, and geometry-aware rendering. PyMuPDF performs direct evidence lookup through the PDF library's standard search interface. BBOX searches an offline page-text and bounding-box index, then renders the retrieved boxes.

\subsection{Evaluation Metrics and Human Verification Protocol}

The automatic metric is quote-level localization rate, i.e., the fraction of quotes for which a method returns at least one candidate page region:
\begin{equation}
\small
\mathrm{LocalizationRate}=
\frac{\#\{\text{quotes with a returned candidate region}\}}
{\#\{\text{evaluated quotes}\}}.
\end{equation}
We separately assess visual correctness on 18,572 reviewed quotes per method, with verification completed by six evaluators. Manual percentages use this reviewed cohort as their denominator; the remaining eight quotes have no manual labels. Each quote--method output receives one of three evaluation labels: \emph{Correct} (the intended evidence is fully and accurately highlighted), \emph{Incorrect} (a localization is returned but is not fully accurate, including minor boundary excess or omission), or \emph{Not found} (no usable localization). Each item has one final recorded label. Automatic localization rate measures candidate availability, whereas these manual labels assess visual correctness.

\subsection{Overall Performance}

\begin{figure}[t]
\centering
\includegraphics[width=\columnwidth]{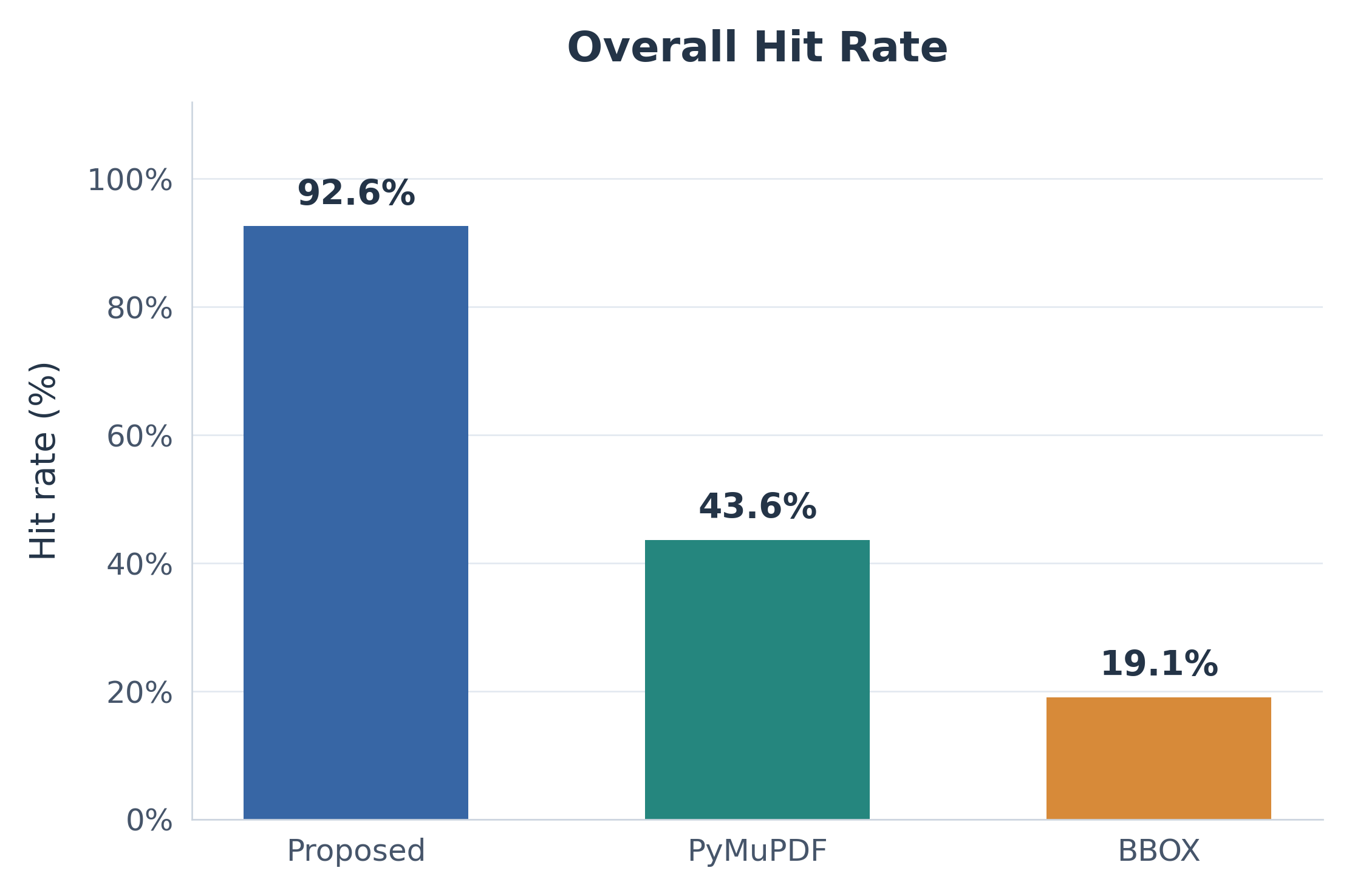}
\caption{Quote-level automatic localization rate.}
\label{fig:overall}
\end{figure}

Table~\ref{tab:summary} and Figure~\ref{fig:overall} show that the proposed method achieves a 92.6\% automatic localization rate, versus 43.6\% for text search and 19.1\% for precomputed BBOX. Reliable localization therefore requires matching positions, source characters, and page geometry in a shared representation.

\subsection{Component Ablation}
\begin{figure}[t]
  \centering
  \includegraphics[width=\columnwidth]{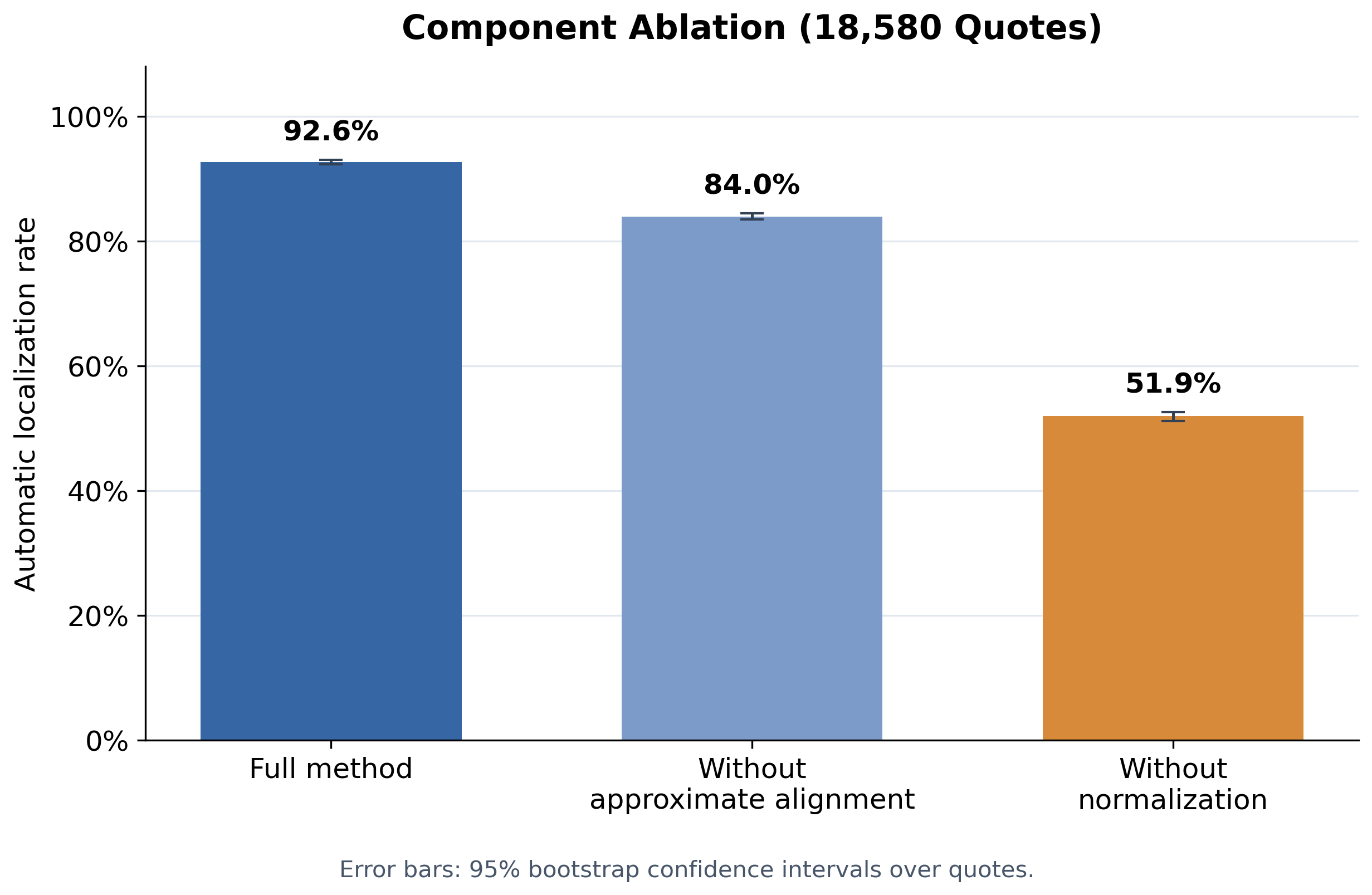}
  \caption{Component ablation on the full evaluation corpus. Error bars show 95\% bootstrap confidence intervals over quotes.}
  \label{fig:ablation}
\end{figure}

Figure~\ref{fig:ablation} isolates the contributions of the proposed normalization and approximate alignment on the same 1,020 papers and 18,580 quotes. Removing approximate alignment disables the LCS fallback while retaining normalized exact matching. Removing the proposed normalization instead uses the PDF.js-extracted text stream directly while retaining the same LCS fallback. This comparison evaluates the added benefit of our normalization beyond the extracted PDF.js text representation. The results show that the proposed normalization provides the larger gain in localization rate, while approximate alignment further recovers disrupted token sequences.

We further conduct a rendering-only boundary ablation on the 17,213 successfully localized quotes: holding the matched source spans fixed, we compare character-range clipping with highlighting every matched PDF.js text item. Whole-item rendering adds 62.8 source characters per quote on average (26.8\% of highlighted characters), whereas clipping introduces no whole-item excess by construction; thus, provenance mapping improves highlight specificity without changing localization decisions.

\subsection{Robustness to Evidence Length}

\begin{figure}[t]
\centering
\includegraphics[width=\columnwidth]{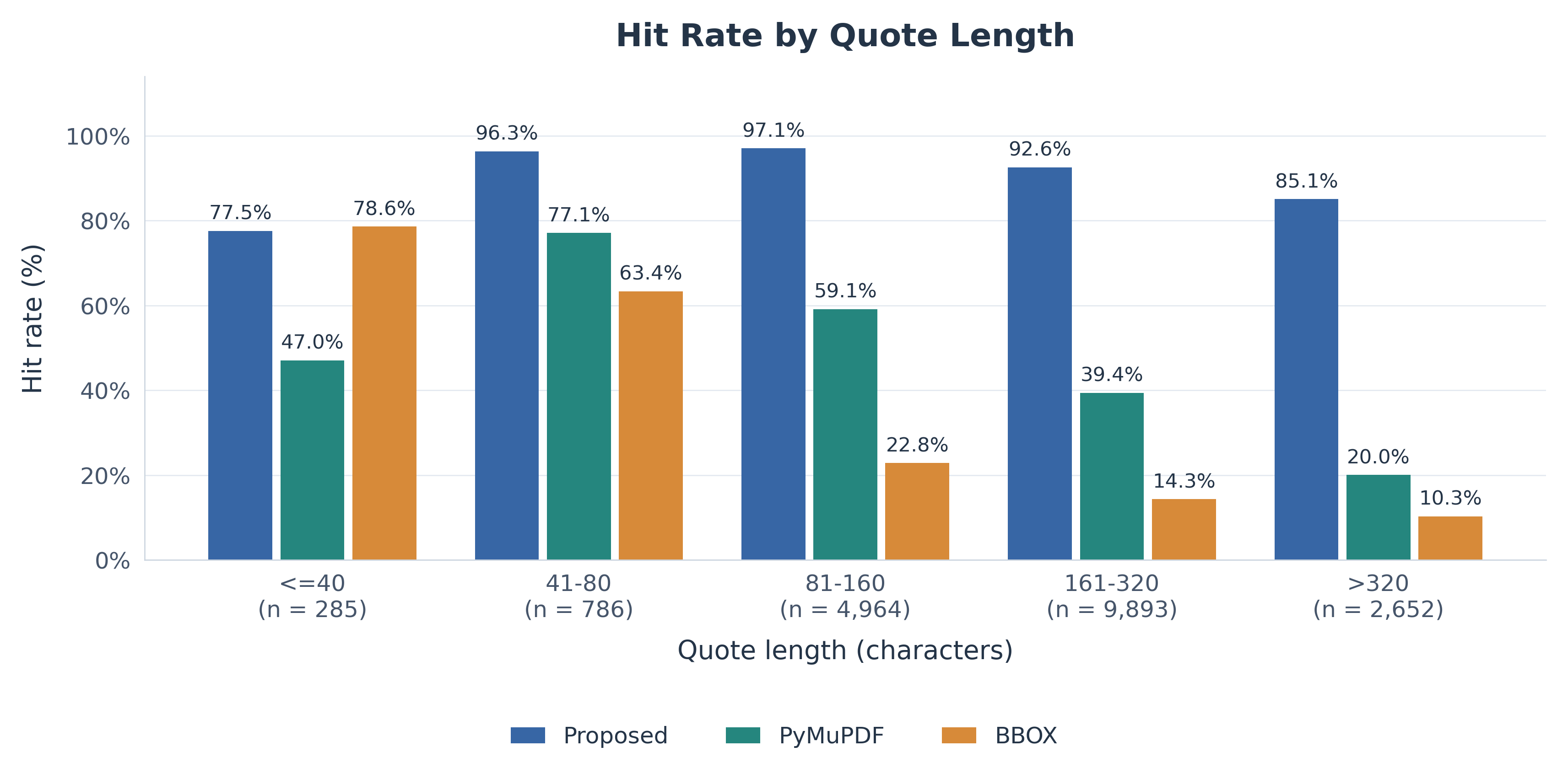}
\caption{Quote-level automatic localization rate by evidence length.}
\label{fig:length}
\end{figure}

Figure~\ref{fig:length} compares localization rates across quote-length groups. The proposed method peaks at 81--160 characters and then declines, but remains substantially ahead of both baselines for longer quotes. BBOX slightly exceeds the proposed method in the shortest group. PyMuPDF peaks at 41--80 characters before declining, whereas BBOX declines across all groups. Longer strings can cross text items and line breaks or contain marker insertions and encoding variation; provenance mapping recovers supported text without expanding highlights to unmatched content.

\subsection{Human Evaluation}

Six evaluators from our laboratory, including faculty members and senior students, manually assessed the localization outputs. Each quote--method output was randomly assigned to one evaluator, who compared the highlight with the corresponding evidence in the original PDF. A result was labeled \emph{Correct} only when both its location and boundaries matched exactly; any positional error, excess, or omission was labeled \emph{Incorrect}, and no usable localization was labeled \emph{Not found}.

\begin{figure}[t]
\centering
\includegraphics[width=\columnwidth]{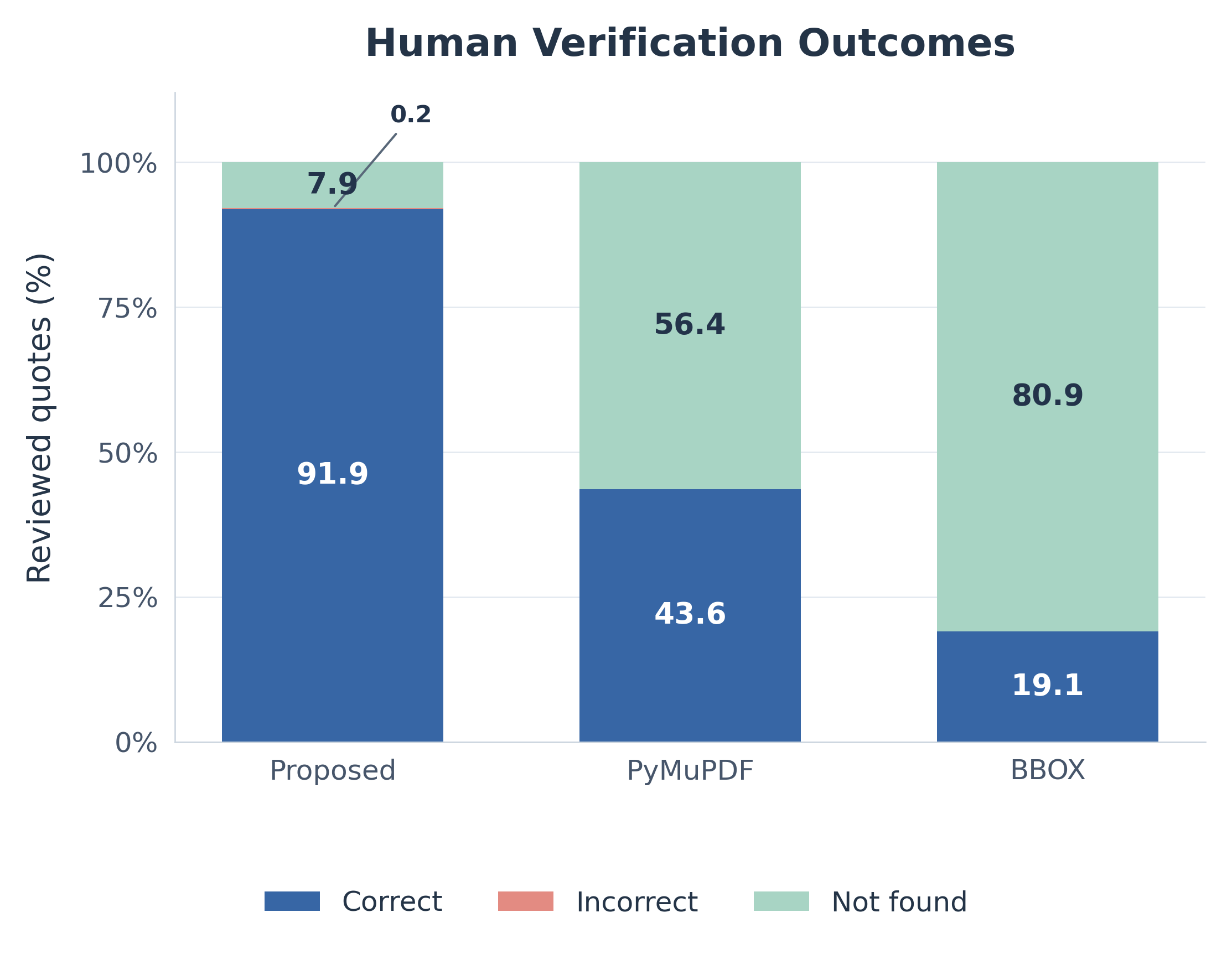}
\caption{Manual verification on 18,580 quotes per method. Correct requires fully accurate localization; partial or inaccurate localizations are Incorrect.}
\label{fig:human}
\end{figure}

Figure~\ref{fig:human} summarizes the three-category results. The proposed method achieves the highest Correct rate, while both baselines have larger Not found shares. Table~\ref{tab:summary} reports the manual and automatic localization rates using their respective evaluation cohorts.

\begin{table}[t]
\centering
\caption{Quote-level results. Loc. Rate uses all 18,580 quotes; Accuracy and Not Found Rate are the manual Correct and Not found shares among 18,572 reviewed quotes. All rates are rounded to one decimal place.}
\label{tab:summary}
\setlength{\tabcolsep}{3pt}
\small
\begin{tabular}{@{}lccc@{}}
\toprule
Method & Loc. Rate & Accuracy & Not Found Rate \\
\midrule
Proposed & 92.1\% & 91.9\% & 7.9\% \\
Text-search baseline & 43.6\% & 43.6\% & 56.4\% \\
Precomputed-BBOX baseline & 19.1\% & 19.1\% & 80.9\% \\
\bottomrule
\end{tabular}
\end{table}

\subsection{Error Analysis}
\label{sec:error-analysis}

Remaining errors mainly arise from complex layouts that disrupt extraction order, proportional widths that approximate variable character widths, and severe extraction errors beyond the normalization rules and LCS threshold.

The framework assumes a usable PDF text layer; separating matching from rendering makes residual errors easier to diagnose.

The principal qualitative advantage is boundary control. If an evidence string ends at ``activities'' but its item continues with ``It is worth noting that'', whole-item highlighting includes unsupported text. Our provenance mapping crops the rectangle at the final evidence character; token alignment likewise excludes inserted markers.

\section{Limitations and Future Work}
\label{sec:limitations-future-work}

The current implementation estimates character geometry from item width, so variable-width fonts can yield loose boundaries. Glyph-level metrics or native text-layer positioning could address this limitation. Layout-aware analysis could further improve multi-column reading order~\cite{huang2022layoutlmv3,shen2021layoutparser}, while confidence calibration and lightweight correction could support large-scale verification.

\section{Conclusion}
\label{sec:conclusion}

We present a source-preserving framework that maps normalized evidence to source characters and page geometry for robust matching and precise highlights. Experiments show substantial improvements over text-search and precomputed-BBOX baselines, especially for long evidence strings. The ablation results further show that both normalization and approximate token alignment are necessary for this gain.

\clearpage

\noindent
\textbf{Acknowledgments.}
This work was supported by Strategic Priority ResearchProgram of Chinese Academy of Sciences (XDA0490000)
The authors have no relevant financial or nonfinancial interests to disclose.

\noindent\textbf{Compliance with Ethical Standards}
This study does not involve human participants or animals. No ethical approval was required.

\bibliographystyle{IEEEbib}
\bibliography{refs}

\end{document}